\documentclass[conference]{IEEEtran}
\IEEEoverridecommandlockouts
\usepackage{cite}
\usepackage{amsmath,amssymb,amsfonts}
\usepackage{algorithmic}
\usepackage{graphicx}
\usepackage{tikz}
\usepackage{pgfplots}
\pgfplotsset{compat=1.8}

\usepackage{hyperref}

\usepackage{subfigure}
\usepackage{xcolor}

\DeclareMathOperator{\vpc}{VPC}

\newcommand{\E}{\mathbb{E}_{\pi}}
\newcommand{\etal}{et al.\ }
    
\newcommand{\quotes}[1]{``#1''}

\begin{document}

\title{Using State Predictions for Value Regularization in Curiosity Driven Deep Reinforcement Learning}


\author{Gino Brunner, Manuel Fritsche*\thanks{* Main contributor}, Oliver Richter and Roger Wattenhofer**\thanks{** Authors listed in alphabetical order.}\\
 Department of Information Technology and Electrical Engineering\\
        ETH Z{\"u}rich\\
        Switzerland\\
        {\tt\small \{brunnegi,manuelf,richtero,wattenhofer\}@ethz.ch}
}

\maketitle
\begin{abstract}
Learning in sparse reward settings remains a challenge in Reinforcement Learning, which is often addressed by using intrinsic rewards. One promising strategy is inspired by human curiosity, requiring the agent to learn to predict the future. In this paper a curiosity-driven agent is extended to use these predictions directly for training. To achieve this, the agent predicts the value function of the next state at any point in time. Subsequently, the consistency of this prediction with the current value function is measured, which is then used as a regularization term in the loss function of the algorithm. Experiments were made on grid-world environments as well as on a 3D navigation task, both with sparse rewards. In the first case the extended agent is able to learn significantly faster than the baselines. 
\end{abstract}

\begin{IEEEkeywords}
curiosity, deep reinforcement learning, intrinsic rewards, machine learning
\end{IEEEkeywords}

\section{Introduction}


In a classical Reinforcement Learning problem an agent is trained to fulfill one or multiple goals in an environment. This is achieved by rewarding the agent whenever it does something desirable. However, in many settings the rewards are extremely rare. In these cases it can be hard for an agent to learn because potential rewards are too far in the future. 
To solve this problem, additional auxiliary tasks which provide ``intrinsic" rewards to the agent have been introduced 
~\cite{unreal}. The purpose of intrinsic rewards is to provide more reward signals to the agent. Auxiliary tasks should be designed such that by solving them, the agent will be able to get higher \quotes{extrinsic} reward, i.e., get better at solving the main objective.
Human curiosity has inspired several of these intrinsic rewards~\cite{noreward-rl, wsmodel, bellmanlike}: The idea is to make predictions about the consequences of an action, i.e., the future state of the environment. The difference between these predictions and the actual consequences are then used as a measure of surprise. 
If a prediction was inaccurate, the agent is surprised and gets \quotes{curious} about it. The farther off the predictions, the higher the intrinsic rewards.
This approach can help an agent to learn (cf. \cite{noreward-rl, wsmodel}), but it often introduces additional neural networks that have to be trained in order to make predictions.  This can make learning unstable due to the difficulty of simultaneously optimizing multiple inter-dependent neural networks.
Therefore, reusing neural network modules for related tasks might help to regularize the training.
Intuitively, if future states can be predicted well, then these predictions should contain valuable information for the choice of the next action. As humans, we often base our actions on predictions, e.g., we bring along an umbrella because we predict that it will rain. 
Existing deep reinforcement learning algorithms like A3C~\cite{a3c} estimate the value function, which is the expected sum of discounted future rewards for a given state and policy. Thus, the value function includes a forecast of future rewards. This forecast should be consistent with the predicted future state for the actions that are taken. If not, then either the state prediction or the forecast must be wrong. 
This idea motivates the algorithm that is proposed in this paper. Building on the A3C algorithm and the work of Pathak \etal\cite{noreward-rl}, we add a regularization term to the loss function to improve the estimate of the value function during training. This regularization term penalizes inconsistencies between the predicted consequences of an action (future state) and the value function. The algorithm is tested on 2 different grid-world mazes and the VizDoom environment~\cite{vizdoom}. The experimental results suggest that penalizing these prediction inconsistencies can improve the performance of an algorithm.\footnote{Our code can be found here: \url{https://github.com/ManuelFritsche/vpc}}

\section{Related Work}
Several different ways of using intrinsic rewards have been proposed: Jaderberg \etal\cite{unreal} introduce multiple pseudo-reward functions that the agent tries to maximize in addition to the extrinsic rewards. 
Their value function replay auxiliary task is similar to the regularization term proposed in this paper. However, our approach does not require a replay buffer and therefore has lower memory requirements.
Houthooft \etal\cite{vime} propose an exploration bonus based on information gain. Sukhbaatar \etal\cite{alicebob} use two versions of the same agent in an adversarial fashion with each agent repeatedly proposing tasks that the other agent is supposed to complete. Some work, e.g., \cite{count1, count2, count3}, has used state visitation counts as a measure of novelty, which is then used to define intrinsic rewards. A review of earlier work on intrinsic rewards can be found in \cite{earlyreview}.
Curiosity-driven exploration is a popular way of defining intrinsic rewards. Schmidhuber \cite{firstcuriosity} proposes an agent with an additional prediction model to obtain a measure of curiosity. Pathak \etal\cite{noreward-rl} extend this model to filter out irrelevant information of the agent's input. Haber \etal\cite{wsmodel} use an adverserial network to further improve this prediction model. Abril \etal\cite{bellmanlike} use curiosity driven exploration with an additional drive to act according to familiar patterns. The resulting agents have a drive to explore regions that are difficult to predict, while they are simultaneously improving their predictions. However, since an additional model is introduced that has to be trained, the computational complexity increases. This work uses a similar method to measure curiosity, but additionally uses the prediction model to improve the policy directly. To the best of our knowledge, non of the related work explored this so far.

 
\section{Curiosity in Reinforcement Learning}
\label{intrinsic_rewards}

In this work a standard reinforcement learning setting is assumed, where an agent interacts with an environment at discrete time steps. At every time step $t$ the agent is in some observable state $s_t$ and chooses an action $a_t$. Depending on $s_t$ and $a_t$ the agent obtains a reward $r_t$ and switches to state $s_{t+1}$ at time $t+1$. The agent acts according to a policy $\pi(a|s)$, which is a probability distribution over the discrete action space given the state $s$. The return $R_t$ at time $t$ is defined as the sum of discounted future rewards: $R_t=\sum_{k=0}^{\infty} \gamma^k r_{t+k}$, where $\gamma \in (0, 1]$ is called the discount factor. This return determines the value of a state, which is measured by the value function $V^{\pi}(s) = \E[R_t |s_t=s]$ of policy $\pi$.

This work builds on the popular A3C algorithm \cite{a3c}, which is a reinforcement learning algorithm that learns a policy $\pi$ and an estimate of the corresponding value function $V^{\pi}(s)$ of a state $s$. It does so, by training a neural network on-policy, i.e., incorporating experiences directly into the network weights through gradient descent without the use of a replay buffer. 
This is done asynchronously with multiple workers to break the correlation of updates a single worker would induce when training on-policy.
The estimate of the value function is used as a critic to improve the policy (actor). 

To effectively learn in a setting where rewards provided by the environment are sparse, intrinsic rewards have been introduced \cite{formalToIM},
denoted by $r_t^i$ in the following. These rewards are added to the extrinsic rewards $r_t^e$ that are obtained from the environment. The agent only sees the sum of the rewards $r_t = r_t^e + r_t^i$ as the reward for each state action pair. The hope is that by using intrinsic rewards the agent's ability to explore improves, which allows it to find more extrinsic rewards.


In this paper the intrinsic reward is inspired by human curiosity, which can be seen as trying to explore the states and actions that yield results that are surprising. To be surprised in the first place, the algorithm needs to do some sort of prediction. The surprise can then be defined as the difference between reality and the prediction.


\begin{figure}
    \centering
    \includegraphics[width=\linewidth]{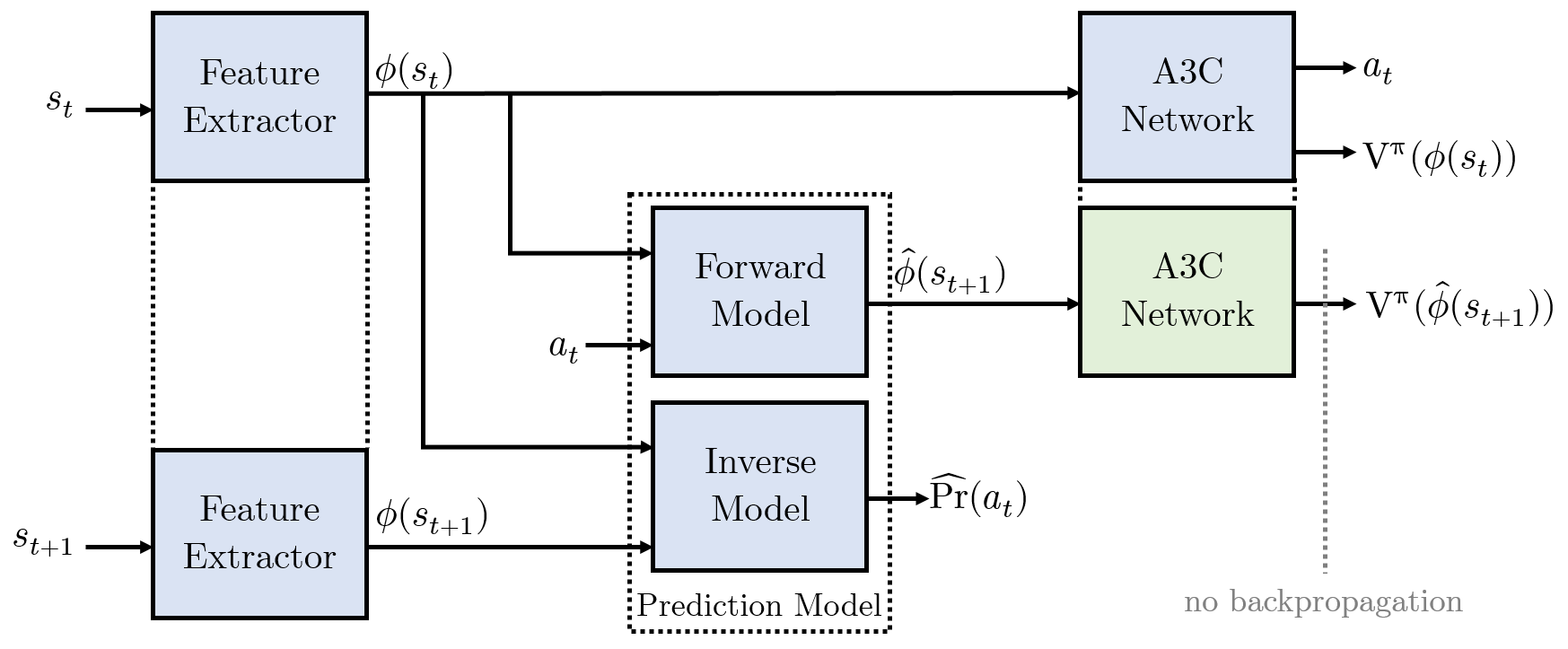} 
    \caption{The A3C architecture with the Forward and Inverse Model (blue part) plus the Value Prediction Consistency addition (green part). All networks use the same Feature Extractor. Networks with the same names, are sharing the same weights (marked with dashed lines), i.e., to calculate $\phi(s_{t+1})$ the same Feature Extractor is used as for $\phi(s_{t})$ and for $V^{\pi}(\widehat{\phi}(s_{t+1}))$ the same network is used as for $V^{\pi}(\phi(s_{t}))$. $V^{\pi}(\widehat{\phi}(s_{t+1}))$ is treated as a constant during training.}
    \label{fig:architecture}
\end{figure}


The architecture described hereafter is shown in Figure~\ref{fig:architecture}. 
Following \cite{noreward-rl}, training a network to predict certain features of the next state works well, which is referred to as \emph{Prediction Model} in the following. This model contains two neural networks that receive their input from the same \emph{Feature Extractor}. This Feature Extractor is also a neural network containing several convolutional layers which learn to extract important features from the raw states $s$. These features are denoted by $\phi(s)$ in the following. The first network that uses $\phi(s)$ as input is called the \emph{Forward Model}, which is trained to output a prediction $\widehat{\phi}(s_{t+1})$ of $\phi(s_{t+1})$, given the state $s_t$ and action $a_t$.
The intrinsic reward is proportional to the prediction error of this Forward Model, with a scaling factor $\beta$:
\begin{equation*}
r_t^i = \beta \left\lVert\widehat{\phi}(s_{t+1}) - \phi(s_{t+1})\right\rVert_2^2
\end{equation*}

The second network that is built on top of the Feature Extractor is called the \emph{Inverse Model}, which produces an estimate $\widehat{\Pr}(a_t)$ of the probability distribution of action $a_t$ that was taken to get from state $s_t$ to state $s_{t+1}$. This model is only used to train the Feature Extractor. Since the network tries to predict the actions that were responsible for a state transition, it has no incentive to learn features that the agent cannot control with its actions. To guarantee that the features used in the A3C Network are consistent with the ones used for the State Prediction Model, all networks use the same Feature Extractor. This is different from \cite{noreward-rl}, where the A3C Network is completely separate from the Forward and Inverse Model.


The Prediction Model is trained with the loss function
\begin{equation*}
L_\text{P} = \frac{\lambda_\text{F}}{2}\left\lVert\widehat{\phi}(s_{t+1}) - \phi(s_{t+1})\right\rVert_2^2 + \lambda_\text{I} H(a_t, \widehat{\Pr}(a_t))
\end{equation*}
where $\lambda_\text{F}$ and $\lambda_\text{I}$ are constants that are used to weight the importance of each part of the loss function. $H(a_t, \widehat{\Pr}(a_t))$ denotes the standard cross entropy function using a one-hot encoding of the true action $a_t$ and the estimated probability distribution $\widehat{\Pr}(a_t))$.


\section{Value Prediction Consistency}
\label{sec:vpc}
So far the described Prediction Model is only used to generate intrinsic rewards. However, as humans we constantly use predictions to plan our next actions. These predictions might not always be correct, but our actions are usually consistent with them. 
In the same way the actions that are taken should be consistent with predictions of the Forward Model.
The A3C algorithm does not explicitly plan the steps it will take in the future, but it assigns a value function $V^\pi(s_t)$ to the states. This value function basically reflects the current plan, because it sums up all the expected discounted future rewards. Thus, this value function should also be consistent with the prediction of the next state. Mathematically the value function can be expressed as follows:

\begin{equation*}
V^\pi(s_t) =  \E\left[\sum\limits_{k=0}^\infty \gamma^k r_{t+k} \right]  =  \E[r_t] + \gamma V^\pi(s_{t+1})
\end{equation*}


Thus, $V^\pi(s_{t+1})$ can be calculated recursively as

\begin{equation*}
V^\pi(s_{t+1}) = \frac{V^\pi(s_t) - \E[r_t]}{\gamma}
\end{equation*}

In the following the value function estimate of a state $s$ with input features $\phi(s)$ is denoted by $V^\pi(\phi(s))$.

At time $t$ the Prediction Model and the A3C Network have experienced the same information. Thus, to be consistent with each other, the value function of the predicted features $V^\pi(\widehat{\phi}(s_{t+1}))$ should be consistent with the value function $V^\pi(\phi(s_{t+1}))$ that is estimated only with information from time $t$. Since this is not generally the case, we define a \emph{Value Prediction Consistency} (VPC) error as follows:

\begin{equation*}
e_{\vpc} = V^\pi\left(\widehat{\phi}(s_{t+1})\right) - \frac{V^\pi(\phi(s_t)) - \E[r_t]}{\gamma}
\end{equation*}

Calculating the expected reward $\E[r_t]$ is usually not feasible, since only one action can be taken at each state. However, when acting on policy $\pi$ the reward $\overline{r}_t$ that is obtained at every step is an unbiased sample of the random variable $r_t$. Thus, $\overline{r}_t$ is a reasonable approximation for $\E[r_t]$. This yields:

\begin{equation*}
e_{\vpc} \approx V^\pi\left(\widehat{\phi}(s_{t+1})\right) - \frac{V^\pi(\phi(s_t)) - \overline{r}_t}{\gamma}
\end{equation*}

This error can now be calculated at every iteration of the A3C algorithm. Reducing $e_{\vpc}$ of the value function estimate increases its consistency with the Prediction Model. An addition to this architecture that is able to calculate the components of $e_{\vpc}$ is shown in Figure~\ref{fig:architecture} (green part).

In an environment where the agent encounters hardly any rewards, there is little information to train a neural network to estimate the value function. It is easier to train the Prediction Model than to train the A3C Network in these cases, because the Prediction Model may gain additional information with every step. Value Prediction Consistency introduces additional information for the A3C Network, which can benefit the training. In practice this can be achieved by using $e_{\vpc}$ as a regularization term in the loss function. Since it is assumed that the Prediction Model trains faster than the A3C Network, it makes sense to backpropagate only through the A3C Network, but not through the Prediction Model (as shown in Figure~\ref{fig:architecture}), such that the A3C network learns from the Prediction Model and not the other way around. Using the constant $\lambda_{\vpc}$ to weight the regularization term, the loss function changes to
\begin{equation*}
L = L_{\text{A3C}} + L_P + L_{\vpc} \text{ with } L_{\vpc}=\lambda_{\vpc}*e_{\vpc} 
\end{equation*}

\section{Experiments}
In this section the algorithm described in Section~\ref{sec:vpc} is evaluated in different deterministic environments and compared to other baseline algorithms. 


\subsection{Algorithms}
\label{sec:algorithms}
In all algorithms the same Feature Extractor architecture is used, which consists of 4 convolutional layers with 32 filters each, a stride of 2 and 3x3 kernels. Between the layers an ELU activation function \cite{elu} is used.

\subsubsection{A3C}
This is the basic implementation of the A3C algorithm \cite{a3c}. The output of the Feature Extractor $\phi(s_t)$ is fed into an LSTM with 256 units. The value function $V^\pi(s_t)$ and the action $a_t$ are then estimated by separate fully connected layers that use the output of the LSTM units as inputs. 

\subsubsection{PRED (ours)}
Additional to the A3C architecture, the Prediction Model is used as described in Section~\ref{intrinsic_rewards} and Figure~\ref{fig:architecture} (blue part). The Forward and Inverse Model use the same Feature Extractor as the A3C Network uses. For the Forward Model two fully connected layers are used with $\phi(s_t)$ and $a_t$ as input and a ReLU activation function in between. For the Inverse Model the features $\phi(s_t)$ and $\phi(s_{t+1})$ are calculated and then fed into a fully connected layer with a ReLU activation function. On top of this layer another fully connected layer is used after which a softmax is applied to obtain an estimate of the probability distribution of action $a_t$. 

\subsubsection{ICM}
The Internal Curiosity Module (ICM) was proposed by \cite{noreward-rl} and is used for comparison. It is similar to PRED, with the only difference being that the Prediction Model does not share the Feature Extractor with the A3C Network. It uses a duplicate of the Feature Extractor with different weights for the Prediction Model.

\subsubsection{VPC (ours)}
This is the architecture that is described in Section~\ref{sec:vpc} and Figure~\ref{fig:architecture}. After calculating the value function of $s_t$, the features of the prediction $\widehat{\phi}(s_{t+1})$ are fed into the LSTM of the A3C network to obtain $V^{\pi}\left(\widehat{\phi}(s_{t+1})\right)$. To predict $V^{\pi}\left(\widehat{\phi}(s_{t+1})\right)$ the LSTM is set to the state that it has after estimating $V^{\pi}\left(\phi(s_{t})\right)$, i.e.\ first $V^\pi (\phi(s_t))$ is estimated and then $V^{\pi}\left(\widehat{\phi}(s_{t+1})\right)$. For the next step the LSTM is reset to the state it had after calculating $V^\pi (s_t)$. This is done to make sure that the value prediction does not interfere with the A3C Network directly, but only through the loss function. $V^{\pi}\left(\widehat{\phi}(s_{t+1})\right)$ is treated as a constant in the regularization term, as described in Section~\ref{sec:vpc}.


\begin{figure}[t]
    \centering
    \includegraphics[width=0.45\linewidth, height=0.3\columnwidth]{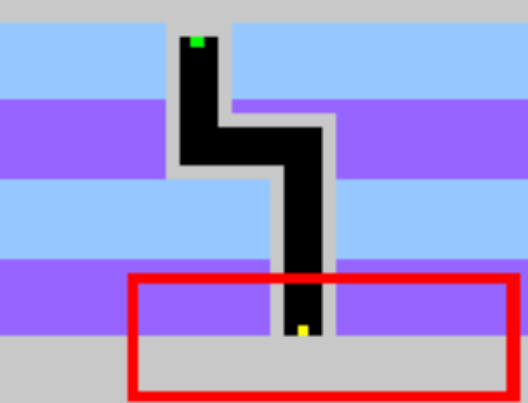}
    \includegraphics[width=0.45\linewidth, height=0.3\columnwidth]{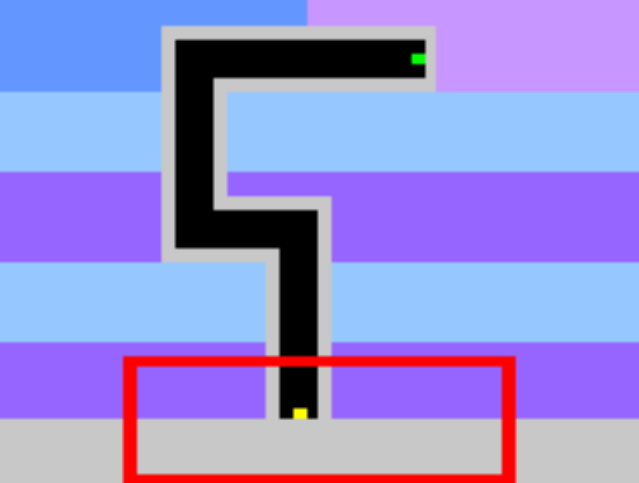} 
    \caption{The Maze A (left) and Maze B (right) environments used for testing. The yellow dot marks the start point and the green dot marks the point where the final reward is obtained. The red frame indicates what the agent sees at the start of each episode. It is only possible to move in the black area.}
    \label{fig:maze}
\end{figure}

\begin{figure}
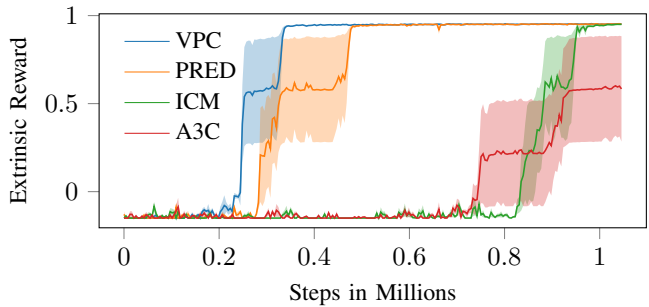

    \centering
    \include{figures/maze150reward}
    \vspace{-1.5em}
    \caption{The average extrinsic reward per episode in the maze A is shown for different architectures. The mean is taken over the average of all workers of three independent training runs with the shaded areas representing the standard error of the mean.}
    \label{fig:results_maze150}
\end{figure}

\subsection{Experiments on Grid World}
\label{sec:grid_world}
One environment that is used for evaluation is a grid world. 
The agent acts in a 2D maze with a top down view of the surroundings. However, it does not see the whole maze, but only a window of $10\times30$ points 
around itself. It has to navigate the maze and arrive at a certain marked spot to obtain a reward of $1$. An episode ends either after a certain amount of steps or if the agent arrives at the final reward. With every step the agent takes it gets an extrinsic reward of $-0.001$, which encourages finding the final reward as quickly as possible.
\subsubsection{Maze A}
\label{sec:small_maze}
The first maze on which the algorithm was evaluated is shown in Figure~\ref{fig:maze} (left). An optimal agent would need $30$ steps to arrive at the final reward. An episode ends after $150$ steps, if the agent does not find the final reward before.
In Figure~\ref{fig:results_maze150} the average extrinsic reward per episode is plotted for the different architectures. In PRED $\lambda_F$ is set to $0.2$ and $\lambda_I$ is chosen as $0.8$, with ICM using the same parameters. Additionally, in VPC $\lambda_{\vpc}$ is set to $0.1$. All architectures are trained with $16$ workers in parallel using Adam~\cite{adam} with a learning rate of $10^{-4}$. The scaling factor for the intrinsic reward $\beta$ is set to $5\cdot 10^{-4}$. The remaining parameters were taken from \cite{noreward-rl}.
One can see that adding curiosity to A3C allows the algorithm to learn much faster in this environment. PRED outperforms ICM, which might be the case because the features that the Prediction Model learns are also good for determining a policy in the A3C Network. Also, it is apparent that adding VPC improves the performance even more and reduces the variance among the training runs. 

\subsubsection{Maze B}
\begin{figure}
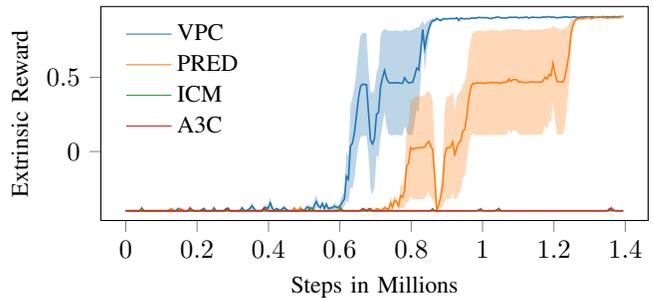

    \centering
    \include{figures/maze400reward}
    \vspace{-1.5em}
    \caption{The average extrinsic reward per episode in the bigger maze B is shown for different architectures. The mean is taken over the average of all workers of three independent training runs with the shaded areas representing the standard error of the mean.}
    \label{fig:results_maze400}
\end{figure}

The second maze that was used for evaluation is shown in Figure~\ref{fig:maze} (right). Here, an optimal agent takes $50$ steps to finish the maze and the episode finishes after a maximum of $400$ steps.
Figure \ref{fig:results_maze400} shows the average extrinsic reward per episode for the different architectures, using the same hyperparameters as for maze A.
The pure A3C agent and the ICM agent do not learn to find the final reward in the tested number of steps. 
One can also see that adding VPC further improves upon PRED by increasing the training speed, as well as reducing the variance among the training runs.

\subsection{Experiments on VizDoom}

\begin{figure}
    \centering
    \includegraphics[width=0.45\linewidth]{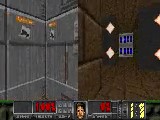} 
    \includegraphics[width=0.45\linewidth]{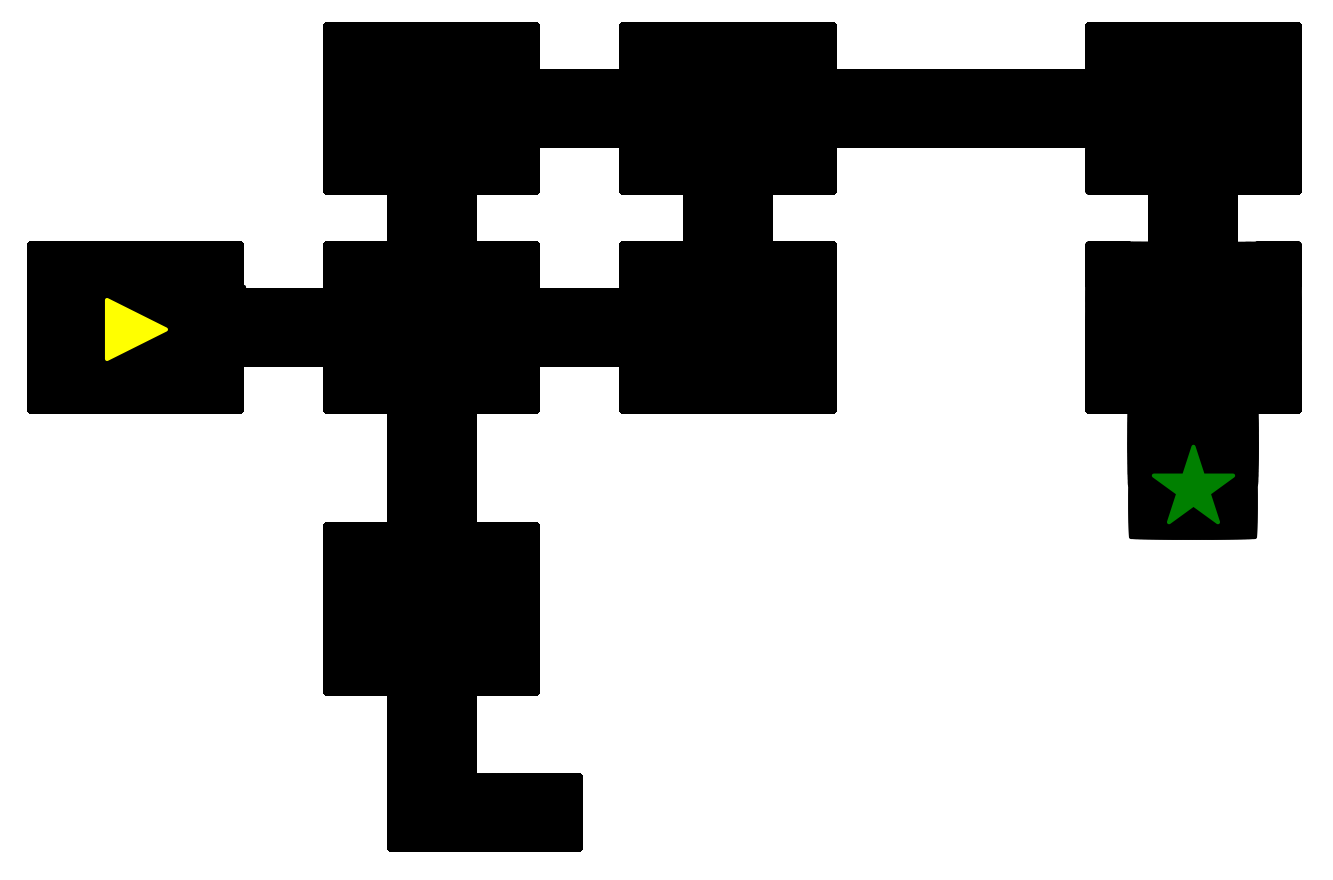} 
    \caption{Left: Unprocessed example frame that the agent gets as input. Right: Scheme of the doom environment showing the map in black. At the beginning of each episode the agent spawns at the yellow triangle facing to the right. It takes at least $270$ steps to arrive at the final reward which is marked with a green star.}
    \label{fig:doom_setup}
\end{figure}

\begin{figure}
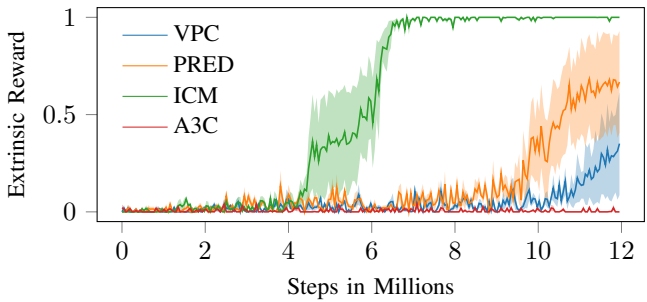

    \centering
    \include{figures/doomSparsereward}
    \vspace{-1.5em}
    \caption{The average extrinsic reward per episode is shown for different architectures. The mean is taken over the average of all workers of three independent training runs with the shaded areas representing the standard error of the mean.}
    \label{fig:results_doom}
\end{figure}

\begin{figure}[t]
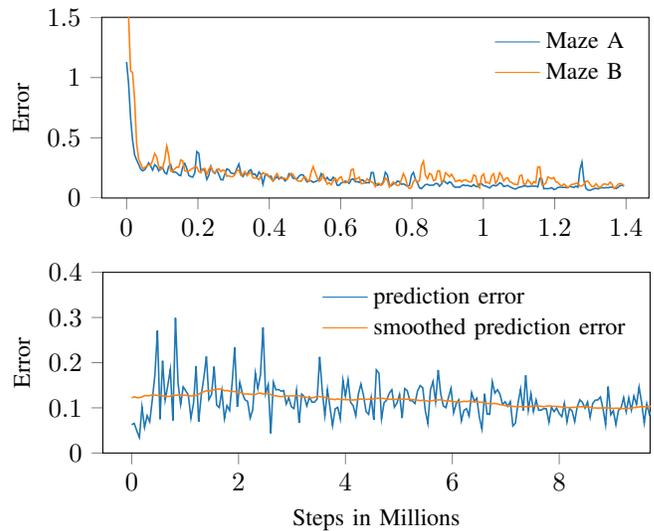

    \centering
    \include{figures/mazeError}
    \vspace{-2em}
    \include{figures/doomError}
    \vspace{-1em}
    \caption{Average prediction error $\left\lVert \widehat{\phi}(s_{t+1}) - \phi(s_{t+1})\right\rVert_2$ of the VPC agent. Top: Maze environments. Bottom: VizDoom environment. Three independent training runs are averaged to generate the plot. 
}
    \label{fig:pred_error}
\end{figure}

Another experiment was conducted on VizDoom \cite{vizdoom}, which is a Doom based 3D environment. The same setup was used as in the ``sparse reward" setting in \cite{noreward-rl}, which is based on ``DoomMyWayHome-v0" from OpenAI gym \cite{mywayhome}. In this environment the agent has to navigate the maze shown in Figure~\ref{fig:doom_setup} to find a vest for which it obtains a reward of $1$. Four actions are possible - move forward, turn left, turn right or no-action, where each action is repeated $4$ times after choosing it. As input the agent gets RGB images with an example shown in Figure~\ref{fig:doom_setup} (left). As in \cite{noreward-rl}, these are converted to $42\times42$ greyscale images and concatenated with the three previous images to emphasize short temporal dependencies. The number of steps in each episode is limited to $2100$, with the agent always spawning in the same room $270$ steps away from the vest. All architectures are trained with $20$ workers in parallel using the Adam Optimizer \cite{adam} with a learning rate of $10^{-4}$. For all algorithms the hyperparameters were taken from \cite{noreward-rl}. ICM uses $\lambda_{\text{F}}=2$ and $\lambda_{\text{I}}=8$, while in PRED $\lambda_{\text{F}}$ and $\lambda_{\text{I}}$ are $0.2$ and $0.8$. VPC uses the same parameters as PRED with $\lambda_{\vpc}=0.1$.
Figure \ref{fig:results_doom} shows the average extrinsic reward per episode for the algorithms described in Section~\ref{sec:algorithms}. In this case, using separate Feature Extractors as input for the A3C Network and the Prediction Model (ICM) allows the algorithm to learn faster than when the Feature Extractor is shared (PRED). This might be because the feature extractor weights can be specialized during training in the ICM agent. Adding VPC does not improve learning in this environment, which might be due to that the environment is hard to predict, or at least the agent does not learn to predict it accurately. In Figure~\ref{fig:pred_error} the prediction error $\left\lVert \widehat{\phi}(s_{t+1}) - \phi(s_{t+1})\right\rVert_2$ of the VPC agent is plotted. One can see that the error does not improve significantly over the course of training in the VizDoom environment as opposed to the maze environments. Thus, if the Prediction Model does not provide accurate predictions, VPC regularization does not improve training.

\section{Conclusion}

The results on the mazes suggest that valuable information is learned by the Prediction Model, which can be utilized for training a policy. In previous work this information was mainly used for generating intrinsic rewards, which has proven to work well in practice. However, 
using state predictions in a regularization term for the loss of the A3C Network seems to be a promising direction for further research.
On our maze environments, the proposed regularization term improves the learning speed and
lowers the variance among training runs. Furthermore, the results show that using the same Feature Extractor for the Prediction Model as for the A3C Network can also make training faster. 
Future research could address a measure of confidence in the predictions to overcome the shortcomings on hard-to-predict environments and/or leverage more information on easy-to-predict environments, e.g., predicting longer roll-outs.

\bibliographystyle{splncs}
\bibliography{references}

\end{document}